\documentclass[10pt,twocolumn,letterpaper]{article}

\usepackage{cvpr}
\usepackage{times}
\usepackage{epsfig}
\usepackage{graphicx}
\usepackage{amsmath}
\usepackage{amssymb}
\usepackage{graphics}
\usepackage{graphicx}
\usepackage[]{graphicx} 
\usepackage{epsfig} 
\usepackage{subfigure}
\usepackage{algorithm}
\usepackage{algorithmic}
\usepackage{stmaryrd}
\usepackage[mathscr]{eucal}
\usepackage{lineno}
\usepackage{color}
\usepackage{filecontents}
\usepackage{subfigure}
\usepackage{multirow}
\usepackage{amsmath}
\usepackage{amssymb}
\usepackage{filecontents}
\usepackage{verbatim} 
\usepackage{tabularx}
\usepackage{lineno}
\usepackage{setspace}
\usepackage{multirow}
\usepackage{textcomp,booktabs}
\usepackage{caption}

\def\X{{\bf X}}
\def\Y{{\bf Y}}

\def\x{{\bf x}}

\def\z{{\bf z}}

\def\U{{\bf U}}

\def\V{{\bf V}}

\def\W{{\bf W}}
\def\w{{\bf w}}
\def\0{{\bf 0}}
\def\1{{\bf 1}}

\def\PsiBM{\boldsymbol{\Psi}}
\def\SigmaBM{\boldsymbol{\Sigma}}

\def\XM{{\mathcal X}}

\def\RB{{\mathbb R}}

\def\eg{\emph{e.g. }}
\def\ie{\emph{i.e. }}

\def\tr{\mathsf{tr}}

\def\etal{{\em et al.\/}\,}

\newtheorem{theorem}{Theorem}

\graphicspath{{./figures/}}   

\usepackage[breaklinks=true,bookmarks=false]{hyperref}

\cvprfinalcopy 

\def\cvprPaperID{****} 

\begin{document}

\title{Low-rank Bilinear Pooling for Fine-Grained Classification}

\author{Shu Kong, Charless Fowlkes\\
Dept. of Computer Science\\
University of California, Irvine\\
{\tt\small \{skong2, fowlkes\}@ics.uci.edu}\\
}

\maketitle

\begin{abstract}
Pooling second-order local feature statistics to form a high-dimensional
bilinear feature has been shown to achieve state-of-the-art performance on a
variety of fine-grained classification tasks.  To address the computational
demands of high feature dimensionality, we propose to represent the covariance
features as a matrix and apply a low-rank bilinear classifier.  The resulting
classifier can be evaluated without explicitly computing the bilinear feature
map which allows for a large reduction in the compute time as well as
decreasing the effective number of parameters to be learned.

To further compress the model, we propose classifier co-decomposition that
factorizes the collection of bilinear classifiers into a common factor and
compact per-class terms.  The co-decomposition idea can be deployed through two
convolutional layers and trained in an end-to-end architecture.  We suggest a
simple yet effective initialization that avoids explicitly first training and
factorizing the larger bilinear classifiers.  Through extensive experiments, we
show that our model achieves state-of-the-art performance on several public
datasets for fine-grained classification trained with only category labels.
Importantly, our final model is an order of magnitude smaller than the recently
proposed compact bilinear model~\cite{gao2015compact}, and three orders smaller
than the standard bilinear CNN model~\cite{lin2015bilinear}.

\end{abstract}

\section{Introduction and Related Work}

\begin{figure*}[t]
\centering
   \includegraphics[width=1\linewidth]{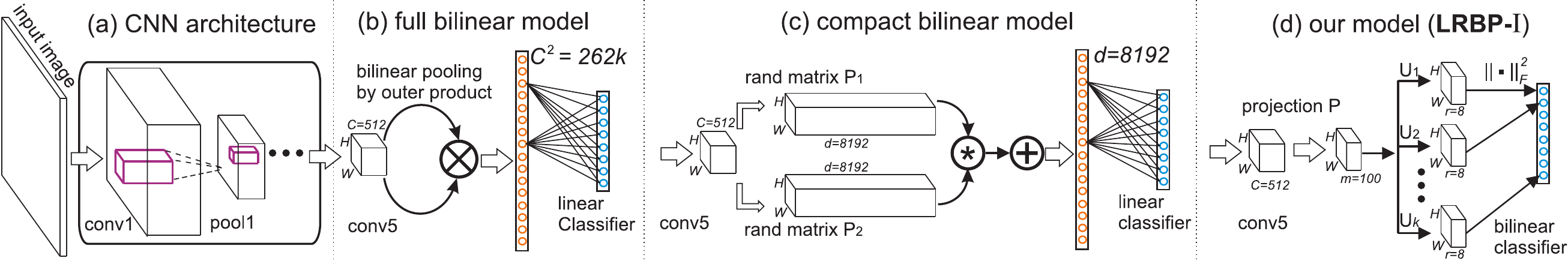}
   \caption{
   We explore models that perform classification using second order statistics
   of a convolutional feature map (a) as input (e.g., VGG16 layer $conv5\_3$).
   Architecture of (b) full bilinear
   model~\cite{lin2015bilinear}, (c) recently proposed compact bilinear
   model~\cite{gao2015compact},  and (d) our proposed low-rank bilinear pooling model (LRBP).
   Our model captures second order statistics
   without explicitly computing the pooled bilinear feature, instead using a
   bilinear classifier that uses the Frobenius norm as the classification
   score.  A variant of our architecture that exploits co-decomposition and computes low-dimensional bilinear features is
   sketched in Figure~\ref{fig:our2_config}.}
\label{fig:brief_pipleline}
\end{figure*}

Fine-grained categorization aims to distinguish subordinate categories within
an entry-level category, such as identifying the bird species or particular
models of aircraft.  Compared to general purpose visual categorization
problems, fine-grained recognition focuses on the characteristic challenge of
making subtle distinctions (low inter-class variance) despite highly variable
appearance due to factors such as deformable object pose (high intra-class
variance). Fine-grained categorization is often made even more challenging
by factors such as large number of categories and the lack of training data.

One approach to dealing with such nuisance parameters has been to exploit
strong supervision, such as detailed part-level, keypoint-level and attribute
annotations~\cite{zhang2015fine,huang2015part,zhang2016}.  These methods learn
to localize semantic parts or keypoints and extract corresponding features
which are used as a holistic representation for final classification.  Strong
supervision with part annotations has been shown to significantly improve the
fine-grained recognition accuracy.  However, such supervised annotations are
costly to obtain.

To alleviate the costly collection of part annotations, some have proposed to
utilize interactive learning~\cite{cui2015fine}.  Partially supervised
discovery of discriminative parts from category labels is also a compelling
approach~\cite{kong2016spatially}, especially given the effectiveness of training with web-scale
datasets~\cite{krause2015unreasonable}.  One approach to unsupervised part
discovery~\cite{simonyan2013deep,simon2014part} uses saliency maps, leveraging
the observation that sparse deep CNN feature activations often correspond to
semantically meaningful regions~\cite{zeiler2014visualizing,liu2015treasure}.
Another recent approach~\cite{wang2016mining} selects parts from a pool of
patch candidates by searching over patch triplets, but relies heavily on
training images being aligned w.r.t the object pose.  Spatial transformer
networks~\cite{jaderberg2015spatial} are a very general formulation that
explicitly model latent transformations that align feature maps prior to
classification. They can be trained end-to-end using only classification loss
and have achieved state-of-the-art performance on the very challenging CUB bird
dataset~\cite{wah2011caltech}, but the resulting models are large and stable
optimization is non-trivial.

Recently, a surprisingly simple method called bilinear
pooling~\cite{lin2015bilinear} has achieved state-of-the-art performance on a
variety of fine-grained classification problems.  Bilinear pooling collects
second-order statistics of local features over a whole image to form holistic
representation for classification.  Second-order or higher-order statistics
have been explored in a number of vision tasks (see e.g.
\cite{carreira2012semantic,koniuszsparse}). In the context of fine-grained
recognition, spatial pooling introduces invariance to deformations while
second-order statistics maintain selectivity.

However, the representational power of bilinear features comes at the cost of
very high-dimensional feature representations (see
Figure~\ref{fig:brief_pipleline} (b)), which induce substantial computational
burdens and require large quantities of training data to fit.  To reduce the
model size, Gao \etal ~\cite{gao2015compact} proposed using compact models
based on either random Maclaurin~\cite{kar2012random} or tensor
sketch~\cite{pham2013fast}. These methods approximate the classifier applied to
bilinear pooled feature by the Hadamard product of projected local features with a
large random matrix (Figure~\ref{fig:brief_pipleline} (c)).  These compact
models maintain similar performance to the full bilinear feature with a $90\%$
reduction in the number of learned parameters.

The original bilinear pooling work of Lin \etal and the compact models of Gao
\etal ignore the algebraic structure of the bilinear feature map; instead they
simply vectorize and apply a linear classifier. Inspired by work on the
bilinear SVM~\cite{pirsiavash2009bilinear}, we instead propose to use a
bilinear classifier applied to the bilinear feature which is more naturally
represented as a (covariance) matrix.  This representation not only preserves
the structural information, but also enables us to impose low-rank constraint
to reduce the degrees of freedom in the parameter to be learned.

Our model uses a symmetric bilinear form so computing the confidence score of
our bilinear classifier amounts to evaluating the squared Frobenius norm of the
projected local features.  We thus term our mechanism maximum Frobenius margin.
This means that, at testing time, we do not need to explicitly compute the
bilinear features, and thus computation time can be greatly reduced under some
circumstances, \eg channel number is larger than spatial size.  We show
empirically this results in improved classification performance, reduces the
model size and accelerates feed-forward computation at test time.

To further compress the model for multi-way classification tasks, we propose a
simple co-decomposition approach to factorize the joint collection of
classifier parameters to obtain a even more compact representation.  This
multilinear co-decomposition can be implemented using two separate linear
convolutional layers, as shown in Figure~\ref{fig:brief_pipleline} (d).  Rather
than first training a set of classifiers and then performing co-decomposition
of the parameters, we suggest a simple yet effective initialization based on
feature map activation statistics which allows for direct end-to-end training.

We show that our final model achieves the state-of-the-art performance on
several public datasets for fine-grained classification by using only the
category label.  It is worth noting that the set of parameters learned in our
model is ten times smaller than the recently proposed compact bilinear
model~\cite{gao2015compact}, and a hundred times smaller than the original full
bilinear CNN model~\cite{lin2015bilinear}.

\section{Bilinear Features Meet Bilinear SVMs}
To compute the bilinear pooled features for an image, we first feed the image
into a convolutional neural network (CNN), as shown in
Figure~\ref{fig:brief_pipleline} (a), and extract feature maps at a specific
layer, say VGG16 $conv5\_3$ after rectification.  We denote the feature map by
$\XM \in \RB^{h\times w \times c}$, where $h$, $w$ and $c$ indicate the height,
width and number of feature channels and denote the feature vector at a
specific location by $\x_i \in \RB^{c}$ where the spatial coordinate index $i
\in [1,hw]$.  For each local feature we compute the outer product, $\x_i\x_i^T$
and sum (pool) the resulting matrices over all $hw$ spatial locations to
produce a holistic representation of the image of dimension $c^2$.  This
computation can be written in matrix notation as $\X\X^T = \sum_{i=1}^{hw}
\x_{i}\x_{i}^T$, where $\X\in \RB^{c\times hw}$ is a matrix by reshaping $\XM$
in terms of the third mode. $\X\X^T$ captures the second-order statistics of
the feature activations and is closely related to the sample covariance matrix.

In the bilinear CNN model~\cite{lin2015bilinear} as depicted in
Figure~\ref{fig:brief_pipleline} (b), the bilinear pooled feature is reshaped
into a vector \\
$\z=vec(\X\X^T)\in\RB^{c^2}$ prior to being fed into a linear classifier
\footnote{Various normalization can be applied here, \eg sign
square root power normalization and $\ell_2$ normalization.  We ignore for now
the normalization notations for presentational brevity, and discuss
normalization in Section~\ref{sec:implementation}. }.

Given $N$ training images, we can learn a linear classifier for a specific
class parameterized by $\w\in\RB^{c^2}$ and $b$.  Denote the bilinear feature
for image-$i$ by $\z_i$ and its binary class label as $y_i=\pm 1$ for
$i=1,\dots,N$.  The standard soft-margin SVM training objective is given by:
\begin{equation}
\begin{split}
\min_{\w,b} \frac{1}{N}\sum_{i=1}^N \max(0, 1-y_i\w^T\z_i+b) + \frac{\lambda}{2}\Vert \w \Vert_2^2
\end{split}
\label{eq:linearSVM}
\end{equation}

\subsection{Maximum Frobenius Margin Classifier}

\begin{figure}[t]
\centering
   \includegraphics[width=1\linewidth]{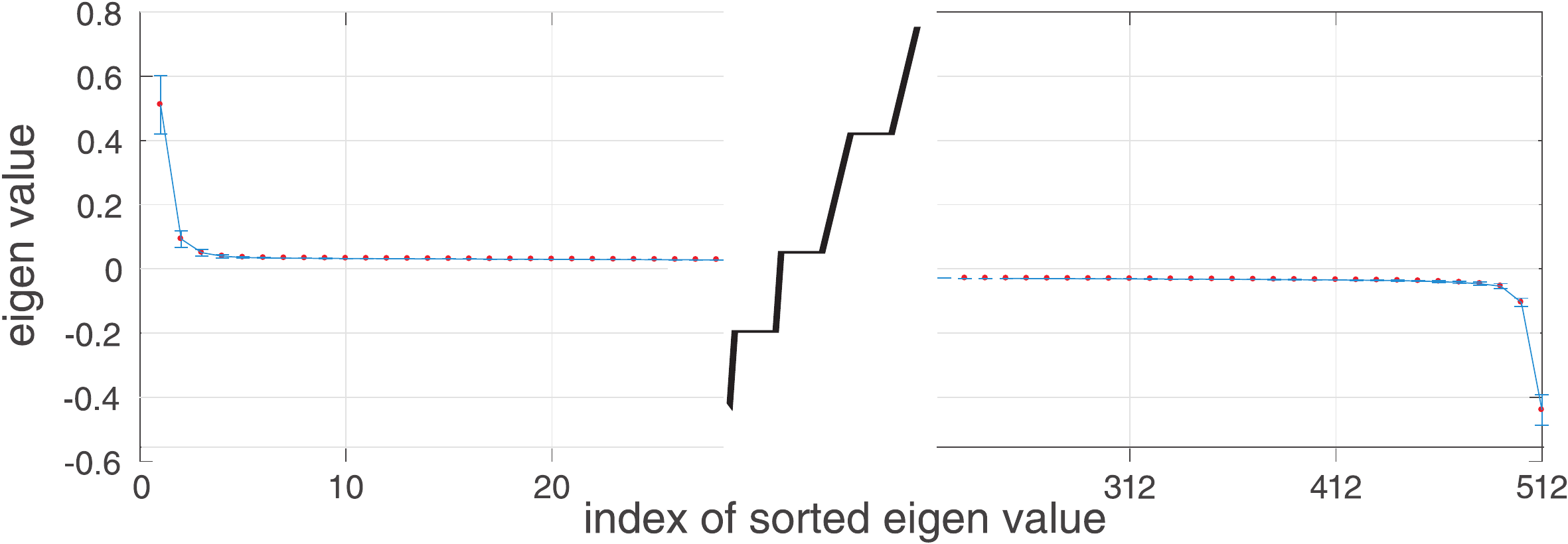}
   \caption{The mean and standard deviation of the eigenvalues the weight
   matrix $\W$ for 200 linear SVM classifiers applied to bilinear features. As
   the plot suggests,
   a large part of the spectrum is typically concentrated
   around 0 with a few large positive and negative eigenvalues. The middle of
   the spectrum is excluded here for clarity. }
\label{fig:eigval_distribution}
\end{figure}

We can write an equivalent objective to Equation~\ref{eq:linearSVM} using the matrix
representation of the bilinear feature as:
\begin{equation}
\begin{split}
\min_{\W,b} \frac{1}{N}\sum_{i=1}^N \max(0, 1-y_i\tr(\W^T \X_i \X_i^T)+b)+ \frac{\lambda}{2}\Vert \W \Vert_F^2 \\
\end{split}
\label{eq:bilinearSVM}
\end{equation}
It is straightforward to show that Equation~\ref{eq:bilinearSVM} is a convex
optimization problem w.r.t. the parameter $\W\in\RB^{c\times c}$ and is
equivalent to the linear SVM.
\begin{theorem}
Let $\w^* \in \RB^{c^2}$ be the optimal solution of the linear SVM in
Equation~\ref{eq:linearSVM} over bilinear features, then $\W^*=mat(\w^*)
\in\RB^{c\times c}$ is the optimal solution in Equation~\ref{eq:bilinearSVM}.
Moreover, $\W^*={\W^*}^T$.
\end{theorem}
To give some intuition about this claim, we write the optimal solution to the
two SVM problems in terms of the Lagrangian dual variables $\alpha$ associated
with each training example:
\begin{equation}
\begin{split}
\w^* = & \sum_{y_i=1} \alpha_i \z_i - \sum_{y_i=-1} \alpha_i \z_i \\
\W^* = & \sum_{y_i=1} \alpha_i \X_i\X_i^T - \sum_{y_i=-1} \alpha_i \X_i\X_i^T \label{eqn:kernelexpansion}\\
&\text{where \ \ } \alpha_i \ge 0, \forall i=1,\dots,N,
\end{split}
\end{equation}
As $\z = vec(\X\X^T)$, it is easy to see that $\w^*=vec(\W^*)$
\footnote{We use $mat(\cdot)$ to denote the inverse of $vec(\cdot)$ so that
$vec(mat(\w))=\w$.}.  Since $\W^*$ is a sum of symmetric matrices, it must also
be symmetric.

From this expansion, it can be seen that $\W^*$ is the difference of two
positive semidefinite matrices corresponding to the positive and negative
training examples.
It is informative to compare Equation
\ref{eqn:kernelexpansion} with the eigen decomposition of $\W^*$
\begin{equation}
\begin{split}
\W^* = & \PsiBM\SigmaBM\PsiBM^T = \PsiBM_+\SigmaBM_+\PsiBM_+^T + \PsiBM_-\SigmaBM_-\PsiBM_-^T \\
    =&\PsiBM_+\SigmaBM_+\PsiBM_+^T - \PsiBM_-\vert \SigmaBM_-\vert \PsiBM_-^T \\
    =&\U_+\U_+^T - \U_-\U_-^T \\
\end{split}
\end{equation}
where $\SigmaBM_+$ and $\SigmaBM_-$ are diagonal matrices containing only
positive and negative eigenvalues, respectively, and $\PsiBM_+$ and $\PsiBM_-$
are the eigenvectors corresponding to those eigenvalues.  Setting
$\U_+=\PsiBM_+\SigmaBM_+^{\frac{1}{2}}$ and $\U_-=\PsiBM_-\vert \SigmaBM_-\vert
^{\frac{1}{2}}$, we have $\W=\U_+\U_+^T - \U_-\U_-^T$.

\begin{figure}[t]
\centering
   \includegraphics[width=1\linewidth]{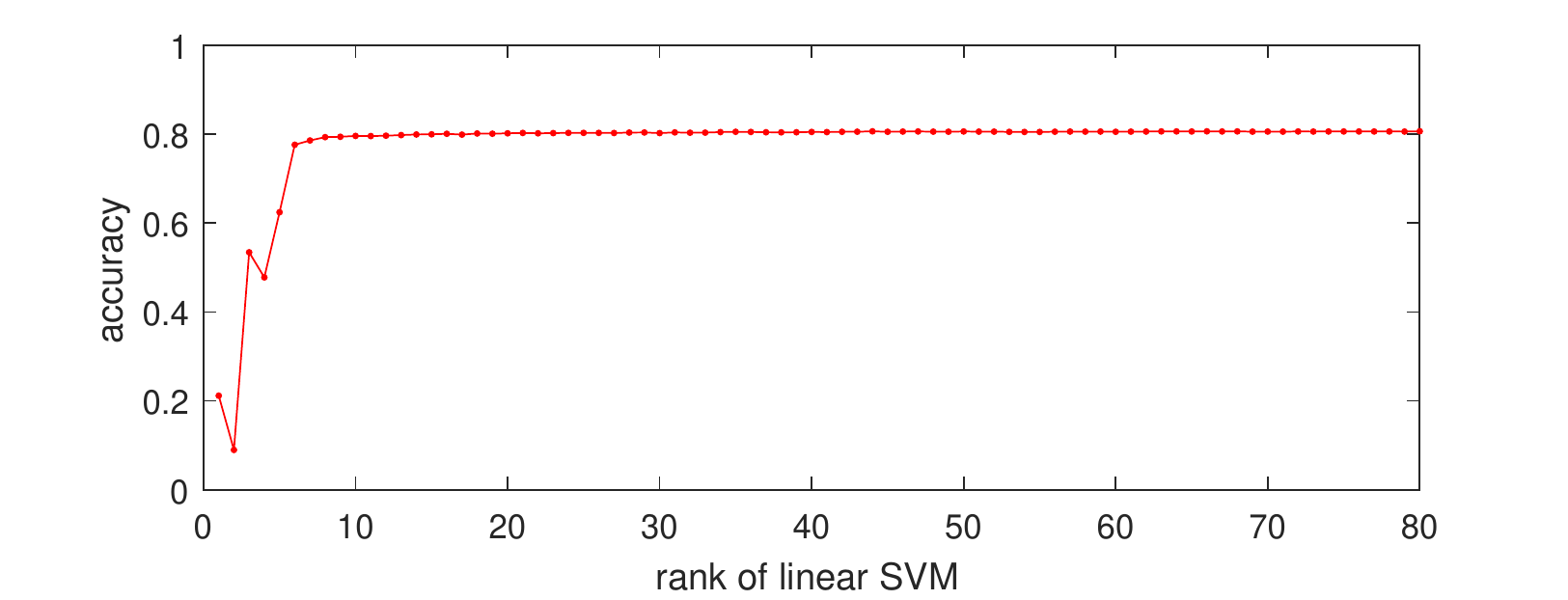}
   \caption{Average accuracy of low-rank linear SVMs.  In this experiment we simply
   use singular value decomposition applied to the set of full rank SVM's for all
   classes to generate low-rank classifiers satisfying a hard rank constraint
   (no fine-tuning). Very low rank classifiers still achieve good performance.}
\label{fig:acc_vs_rank_lsvm}
\end{figure}

In general it will \textit{not} be the case that the positive and negative
components of the eigendecomposition correspond to the dual decomposition
(e.g., that $\U_+\U_+^T = \sum_{y_i=1} \alpha_i \X_i\X_i^T$) since there are
many possible decompositions into a difference of psd matrices. However,
this decomposition motivates the idea that $\W^*$ may well have a good
low-rank decomposition.  In particular we know that $rank(\W^*) < min(N,c)$
so if the amount of training data is small relative to $c$,
$\W^*$ will
necessarily be low rank.
Even with large amounts of training data, SVMs often
produce dual variables $\bf \alpha$ which are sparse so we might expect that the
number of non-zero $\bf \alpha$s is less than $c$.

\paragraph{Low rank parameterization:}
To demonstrate this low-rank hypothesis empirically, we plot in
Figure~\ref{fig:eigval_distribution} the sorted average eigenvalues with
standard deviation of the 200 classifiers trained on bilinear pooled features
from the CUB Bird dataset~\cite{wah2011caltech}.  From the figure, we can
easily observe that a majority of eigenvalues are close to zero and an order
smaller in magnitude than the largest ones.

This motivates us to impose low-rank constraint to reduce the degrees of
freedom in the parameters of the classifier. We use singular value
decomposition to generate a low rank approximation of each of the 200
classifiers, discarding those eigenvectors whose corresponding eigenvalue has
small magnitude.  As shown in Figure~\ref{fig:acc_vs_rank_lsvm}, a rank 10
approximation of the learned classifier achieves nearly the same classification
accuracy as the full rank model. This suggests the set of classifiers can
be represented by $512\times 10 \times 200$ parameters rather than the full set
of $512^2\times 200$ parameters.

\paragraph{Low-rank Hinge Loss:}
In this paper, we directly impose a hard low-rank constraint $rank(\W)=r \ll c$
by using the parameterization in terms of $\U_+$ and $\U_-$, where $\U_+ \in
\RB^{c\times r/2}$ and $\U_- \in \RB^{c\times r/2}$.  This yields the following
(non-convex) learning objective:
\begin{equation}
\small
\begin{split}
 \min_{\substack{\U_+,\U_-,b}} \frac{1}{N} \sum_{i=1}^N
H(\X_i, \U_+, \U_-, b) + \frac{\lambda}{2} R(\U_+, \U_-) \\
\end{split}
\label{eq:splitPosNegObj}
\end{equation}
where $H(\cdot)$ is the hinge loss and $R(\cdot)$ is the regularizer.
The hinge loss can be written as:
\begin{equation}
\small
\begin{split}
H(\X_i, \U_+, \U_-,b) \equiv &  \max(0, 1-y_i\{ \tr( \tilde\W^T \tilde\X ) \}+b)\\
\end{split}
\label{eq:equivObj}
\end{equation}
where
\begin{equation}
\small
\begin{split}
\tilde\W=
\begin{bmatrix}
    \U_+\U_+^T      &\0   \\
    \0       & \U_-\U_-^T  \\
\end{bmatrix},
\tilde \X=
\begin{bmatrix}
    \X_i\X_i^T      &\0   \\
    \0       & -\X_i\X_i^T  \\
\end{bmatrix}.
\end{split}
\label{eq:understandObj}
\end{equation}
While the hinge loss is convex in $\tilde\W$, it is no longer convex in the
parameters $\U_+,\U_-$ we are optimizing.\footnote{Instead of a hard rank
constraint, one could utilize the nuclear norm as a convex regularizer on
$\tilde\W$. However, this wouldn't yield the computational benefits during
training that we highlight here.}

Alternately, we can write the score of the low-rank bilinear classifier as a
difference of matrix norms which yields the following expression of the
hinge-loss:
\begin{equation}
\small
\begin{split}
&H(\X_i, \U_+, \U_-,b)\\
= &  \max(0, 1-y_i\{\tr(\U_+\U_+^T  \X_i \X_i^T) - \tr(\U_-\U_-^T  \X_i \X_i^T) \}+b)\\
= &  \max(0, 1-y_i\{\Vert\U_+^T  \X_i \Vert_F^2 - \Vert \U_-^T  \X_i \Vert_F^2 \}+b)
\label{eq:maximumFroMargin}
\end{split}
\end{equation}
This expression highlights a key advantage of the bilinear classifier, namely
that we never need to explicitly compute the pooled bilinear feature $\X_i
\X_i^T$!

\paragraph{Regularization:}
In the hinge-loss, the parameters $\U_+$ and $\U_-$ are independent of each
other.  However, as noted previously, there exists a decomposition of the
optimal full rank SVM in which the positive and negative subspaces are
orthogonal.  We thus modify the standard $\ell_2$ regularization to include a
positive cross-term $\Vert \U_+^T \U_- \Vert_F^2$ that favors an orthogonal
decomposition.  \footnote{The original $\ell_2$ regularization is given by
$\Vert\W \Vert_F^2 = \Vert \U_+\U_+^T - \U_-\U_-^T \Vert_F^2 = \Vert \U_+\U_+^T
\Vert_F^2 + \Vert \U_-\U_-^T \Vert_F^2 - 2 \Vert \U_+^T \U_-\Vert_F^2$ where
the cross-term actually discourages orthogonality.}
This yields the final objective:
\begin{equation}
\small
\begin{split}
 \min_{\substack{\U_+,b \\ \U_-}} & \frac{1}{N} \sum_{i=1}^N
H(\X_i, \U_+, \U_-, b) \\
+ & \frac{\lambda}{2} (\Vert \U_+\U_+^T\Vert_F^2  + \Vert \U_-\U_-^T \Vert_F^2 + \Vert\U_+^T \U_- \Vert _F^2) \\
\end{split}
\label{eq:finalObj}
\end{equation}

\subsection{Optimization by Gradient Descent}

We call our approach the maximum Frobenius norm SVM.  It is closely related
to the bilinear SVM of Wolf \etal \cite{WolfJhuangHazan2007}, which uses a
bilinear decomposition $\W \approx \U\V^T$. Such non-convex bilinear models
with hard rank constraints are often optimized via alternating descent
\cite{LeeSeung1999,TenenbaumFreeman2000,WolfJhuangHazan2007,pirsiavash2009bilinear}
or fit using convex relaxations based on the nuclear norm \cite{KobayashiIJCV}.
However, our parameterization is actually quadratic in $\U_+,\U_-$ and hence
can't exploit the alternating or cyclic descent approach.

Instead, we optimize the objective function~\ref{eq:finalObj} using stochastic
gradient descent to allow end-to-end training of both the classifier and CNN
feature extractor via standard backpropagation.  As discussed in the
literature, model performance does not appear to suffer from non-convexity
during training and we have no problems finding local minima with good test
accuracy~\cite{dauphin2014identifying,choromanska2015loss}.  The partial
derivatives of our model are straightforward to compute efficiently
\begin{equation}
\small
\begin{split}
\nabla_{\U_+} = & 2\lambda(\U_+\U_+^T\U_+ + \U_-\U_-^T\U_+) \\
&+
\begin{cases}
 0, \ \ \ \ \text{ if } \ \ H(\X_i, \U_+, \U_-,b) \le 0\\
-y_i \X_i \X_i^T\U_+, \  \text{ if } \ \ H(\X_i, \U_+, \U_-,b) > 0
\end{cases}
\\
\nabla_{\U_-} = & 2\lambda(\U_-\U_-^T\U_- + \U_+\U_+^T\U_-) \\
&+
\begin{cases}
 0, \ \ \ \ \text{ if } \ \ H(\X_i, \U_+, \U_-,b) \le 0\\
y_i \X_i \X_i^T \U_-, \  \text{ if } \ \ H(\X_i, \U_+, \U_-,b) > 0
\end{cases}
\\
\nabla_{b}  = &
\begin{cases}
 0, \ \ \ \ \text{ if } \ \ H(\X_i, \U_+, \U_-,b) \le 0\\
-y_i, \  \text{ if } \ \ H(\X_i, \U_+, \U_-,b) > 0
\end{cases}
\end{split}
\label{eq:partialDerivative}
\end{equation}

\section{Classifier Co-Decomposition for Model Compression}
\label{sec:co-decomposition}
In many applications such as fine-grained classification, we are interested in
training a large collection of classifiers and performing k-way classification.
It is reasonable to expect that these classifiers should share some common
structure (e.g., some feature map channels may be more or less informative for
a given k-way classification task).  We thus propose to further reduce the number
of model parameters by performing a co-decomposition over the set of
classifiers in order to isolate shared structure, similar to multi-task
learning frameworks (e.g., \cite{AndoZhangJMLR2005}).

Suppose we have trained $K$ Frobenius norm SVM classifiers for each of $K$
classes.  Denoting the $k^{th}$ classifier parameters as $\U_k=[{\U_+}_k,
{\U_-}_k] \in \RB^{c \times r}$, we consider the following co-decomposition:
\begin{equation}
\small
\begin{split}
\min_{\V_k,{\bf P}} \sum_{k=1}^K \Vert \U_k - {\bf P}\V_k\Vert_F^2,
\end{split}
\label{eq:coDecomp}
\end{equation}
where ${\bf P} \in \RB^{c\times m}$ is a projection matrix that reduces the
feature dimensionality from $c$ to $m<c$,
and $\V_k \in \RB^{m\times r}$ is the new lower-dimensional classifier for the
$k^{th}$ class.

Although there is no unique solution to problem Equation~\ref{eq:coDecomp}, we can
make the following statement
\begin{theorem}
The optimal solution of $\bf P$ to Equation~\ref{eq:coDecomp} spans the subspace of
the singular vectors corresponding to the largest $m$ singular values of
$[\U_1, \dots, \U_K]$.
\end{theorem}
Therefore, without loss of generality, we can add a constraint that $\bf P$ is
a orthogonal matrix without changing the value of the minimum and use SVD on
the full parameters of the $K$ classifiers to obtain ${\bf P}$ and $\V_k$'s.

In practice, we would like to avoid first learning full classifiers $\U_k$ and
then solving for $\bf P$ and $\{\V_k\}$.  Instead, we implement ${\bf P} \in
\RB^{c\times m}$ in our architecture by adding a $1\times 1 \times c\times m$
convolution layer, followed by the new bilinear classifier layer parameterized
by $\V_k$'s. In order to provide a good initialization for $\bf P$,
 we can run
the CNN base architecture on training images and perform PCA on the resulting
feature map activations in order to estimate a good subspace for ${\bf P}$.  We find
this simple initialization of ${\bf P}$ with randomly initialized $\V_k$'s followed
by fine-tuning the whole model achieves state-of-the-art performance.

\begin{table*}[th]
\small
\centering
\caption{A comparison of different compact bilinear models in terms of dimension,
memory, and computational complexity.  The bilinear pooled features are
computed over feature maps of dimension $h \times w \times c$ for a $K$-way
classification problem.  For the VGG16 model on an input image of size $448
\times 448$ we have $h=w=28$ and $c=512$.  The Random Maclaurin and Tensor
Sketch models, which are proposed in~\cite{gao2015compact} based on polynomial
kernel approximation, compute a feature of dimension $d$.  It is shown that
these methods can achieve near-maximum performance with $d=8,192$.  For our
model, we set $m=100$ and $r=8$, corresponding to the reduced feature dimension
and the rank of our low-rank classifier, respectively.  Numbers in brackets
indicate typical values when bilinear pooling is applied after the last
convolutional layer of VGG16 model over the CUB200-2011 bird
dataset~\cite{wah2011caltech} where $K=200$.  Model size only counts the
parameters above the last convolutional layer.
}
\begin{tabular}{l | c c c c c}
\hline
                    & Full Bilinear     & Random Maclaurin  & Tensor Sketch         & LRBP-I               & LRBP-II  \\
\hline
Feature Dim      & $c^2$ [262K]      & $d$ [10K]         & $d$ [10K]             & $mhw$ [78K]        & $m^2$ [10K] \\
\hline
Feature computation    & $O(hwc^2)$        & $O(hwcd)$         & $O(hw(c+d\log d))$    & $O(hwmc)$          & $O(hwmc+hwm^2)$ \\
Classification comp.   & $O(Kc^2)$         & $O(Kd)$           & $O(Kd)$               & $O(Krmhw)$          & $O(Krm^2)$ \\
\hline
Feature Param      & 0                 & $2cd$ [40MB]      & $2c$ [4KB]            & $cm$ [200KB]       & $cm$ [200KB]   \\
Classifier Param      & $Kc^2$ [$K$MB]    & $Kd$ [$K\cdot$32KB]    & $Kd$ [$K\cdot$32KB]        & $Krm$ [$K\cdot$3KB]     & $Krm$ [$K\cdot$3KB] \\
Total ($K=200$)    & $Kc^2$ [200MB]  & $2cd+Kd$ [48MB] & $2c+Kd$ [8MB]              & $cm+Krm$ [0.8MB]          & $cm+Krm$ [0.8MB]  \\
\hline
\end{tabular}
\label{tab:MethodsComparison}
\end{table*}

\begin{figure}[t]
\centering
   \includegraphics[width=1\linewidth]{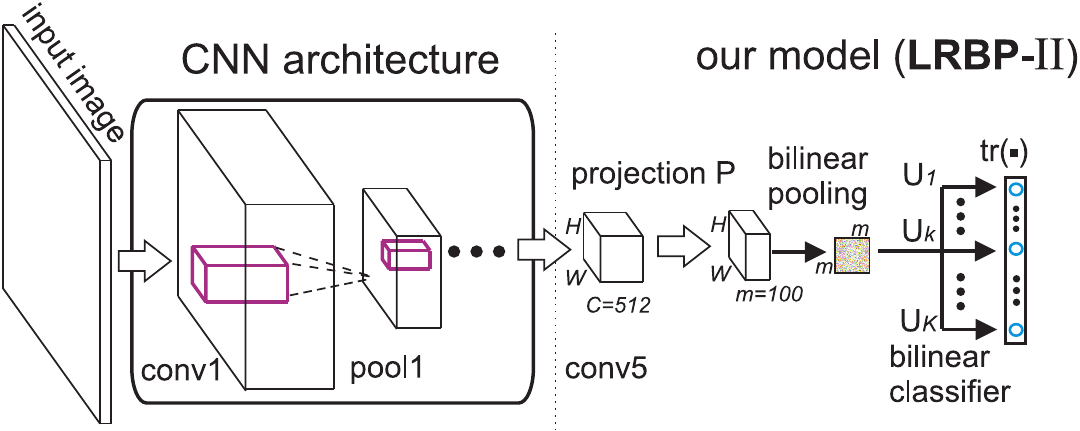}
   \caption{Another configuration of our proposed architecture that explicitly
   computes the bilinear pooling over co-decomposed features of lower
   dimension.}
\label{fig:our2_config}
\end{figure}

\section{Analysis of Computational Efficiency}
\label{sec:complexity}
In this section, we study the computational complexity and model size in
detail, and compare our model to several closely related bilinear methods,
including the full bilinear model~\cite{lin2015bilinear} and two compact
bilinear models~\cite{gao2015compact} by Random Maclaurin and Tensor Sketch.

We consider two variants of our proposed \textit{low-rank bilinear pooling} ({\bf LRBP}) architecture.  In the first, dubbed
\emph{LRBP-I} and depicted in Figure~\ref{fig:brief_pipleline} (d), we use the
Frobenius norm to compute the classification score (see
Equation~\ref{eq:maximumFroMargin}). This approach is preferred when $hw<m$.  In the
second, dubbed \emph{LRBP-II} and depicted in Figure~\ref{fig:our2_config}, we
apply the feature dimensionality reduction using $\bf P$ and then compute the
pooled bilinear feature explicitly and compute the classification score
according to second line of Equation~\ref{eq:maximumFroMargin}.  This has a
computational advantage when $hw>m$.

Table~\ref{tab:MethodsComparison} provides a detailed comparison in terms of
feature dimension, the memory needed to store projection and classifier
parameters, and computational complexity of producing features and classifier
scores.  In particular, we consider this comparison for the CUB200-2011 bird
dataset~\cite{wah2011caltech} which has $K=200$ classes.  A conventional setup
for achieving good performance of the compact bilinear model is that $d=8,192$
as reported in~\cite{gao2015compact}. Our model achieves similar or better
performance using a projection ${\bf P}\in \RB^{512\times 100}$, so that
$m=100$, and using rank $r=8$ for all the classifiers.


From Table~\ref{tab:MethodsComparison}, we can see that Tensor Sketch and our
model are most appealing in terms of model size and computational complexity.
It is worth noting that the size of our model is a hundred times smaller than
the full bilinear model, and ten times smaller than Tensor Sketch.  In
practice, the complexity of computing features in our model $O(hwmc+hwm^2)$ is not
much worse than Tensor Sketch $O(hw(c+d\log(d))$, as $m^2\approx d$, $mc<d\log(d)$ and $m \ll
c$.  Perhaps the only trade-off is the computation in classification step,
which is a bit higher than the compact models.

\section{Experiment Evaluation}
In this section, we provide details of our model implementation along with
description of methods we compare to.
We then investigate design-choices of
our model, \ie the classifier rank and low-dimensional subspace determined by
projection $\bf P$.
Finally we report the results on four commonly used
fine-grained benchmark datasets and describe several methods for generating
qualitative visualizations that provide understanding of image features
driving model performance.

\subsection{Implementation Details}
\label{sec:implementation}
We implemented our classifier layers within matconvnet
toolbox~\cite{vedaldi2015matconvnet} and train using SGD on a single Titan X
GPU. We use the VGG16 model~\cite{simonyan2014very} which is pretrained on
ImageNet, removing the fully connected layers, and inserting a co-decomposition
layer, normalization layer and our bilinear classifiers.  We use PCA to
initialize ${\bf P}$ as described in Section~\ref{sec:co-decomposition}, and
randomly initialize the classifiers.  We initially train only the classifiers,
and then fine-tune the whole network using a batch size of 12 and a small
learning rate of $10^{-3}$, periodically annealed by 0.25, weight decay of
$5\times 10^{-4}$ and momentum 0.9.  The code and trained model will be
released to the public.

We find that proper feature normalization provides a non-trivial improvement in
performance.  Our observation is consistent with the literature on applying
normalization to deal with visual
burstiness~\cite{jegou2009burstiness,lin2015bilinear}.  The full bilinear CNN
and compact bilinear CNN consistently apply sign square root and $\ell_2$
normalization on the bilinear features.  We can apply these normalization
methods for our second configuration (described in
Section~\ref{sec:complexity}).  For our first configuration, we don't
explicitly compute the bilinear feature maps. Instead we find that sign square
root normalization on feature maps at $conv5\_3$ layer results in performance
on par with other bilinear pooling methods while additional $\ell_2$
normalization harms the performance.

\begin{figure*}[t]
    \centering
    \begin{minipage}{0.33\textwidth}
        \centering
        \includegraphics[width=1\linewidth]{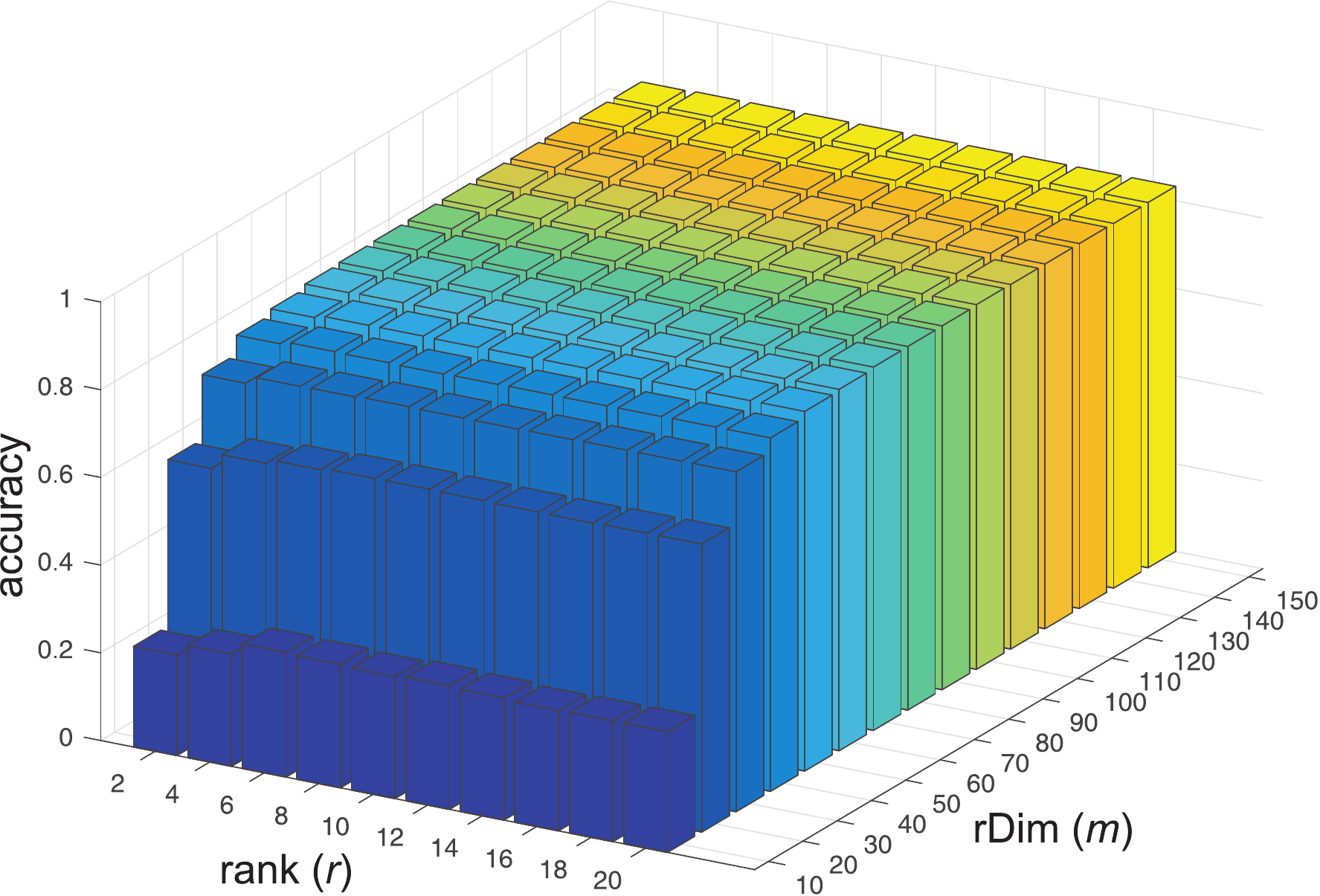}
        \captionsetup{width=0.94\textwidth}
        \caption{Classification accuracy on CUB-200 dataset~\cite{wah2011caltech} vs. reduced dimension ($m$) and rank ($r$).}
        \label{fig:acc_dimRed_rank}
    \end{minipage}
    \begin{minipage}{.33\textwidth}
        \centering
        \includegraphics[width=1\linewidth]{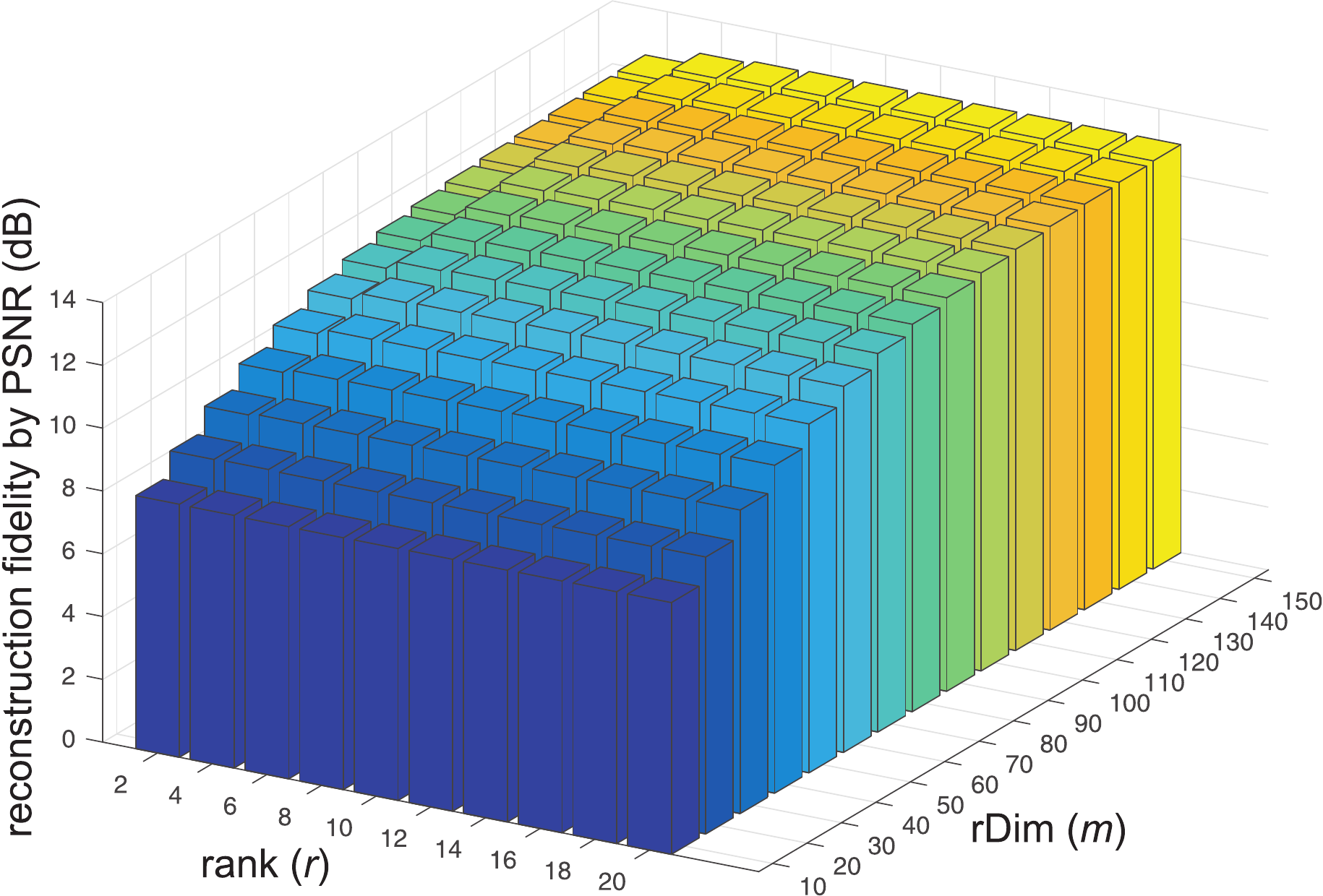}
        \captionsetup{width=0.9\textwidth}
        \caption{Reconstruction fidelity of classifier parameters measured by peak signal-to-noise ratio versus reduced dimension ($m$) and rank (r).
        }
        \label{fig:reconErr_vs_rDimRank_lsvm}
    \end{minipage}%
    \begin{minipage}{0.33\textwidth}
        \centering
        \includegraphics[width=1\linewidth]{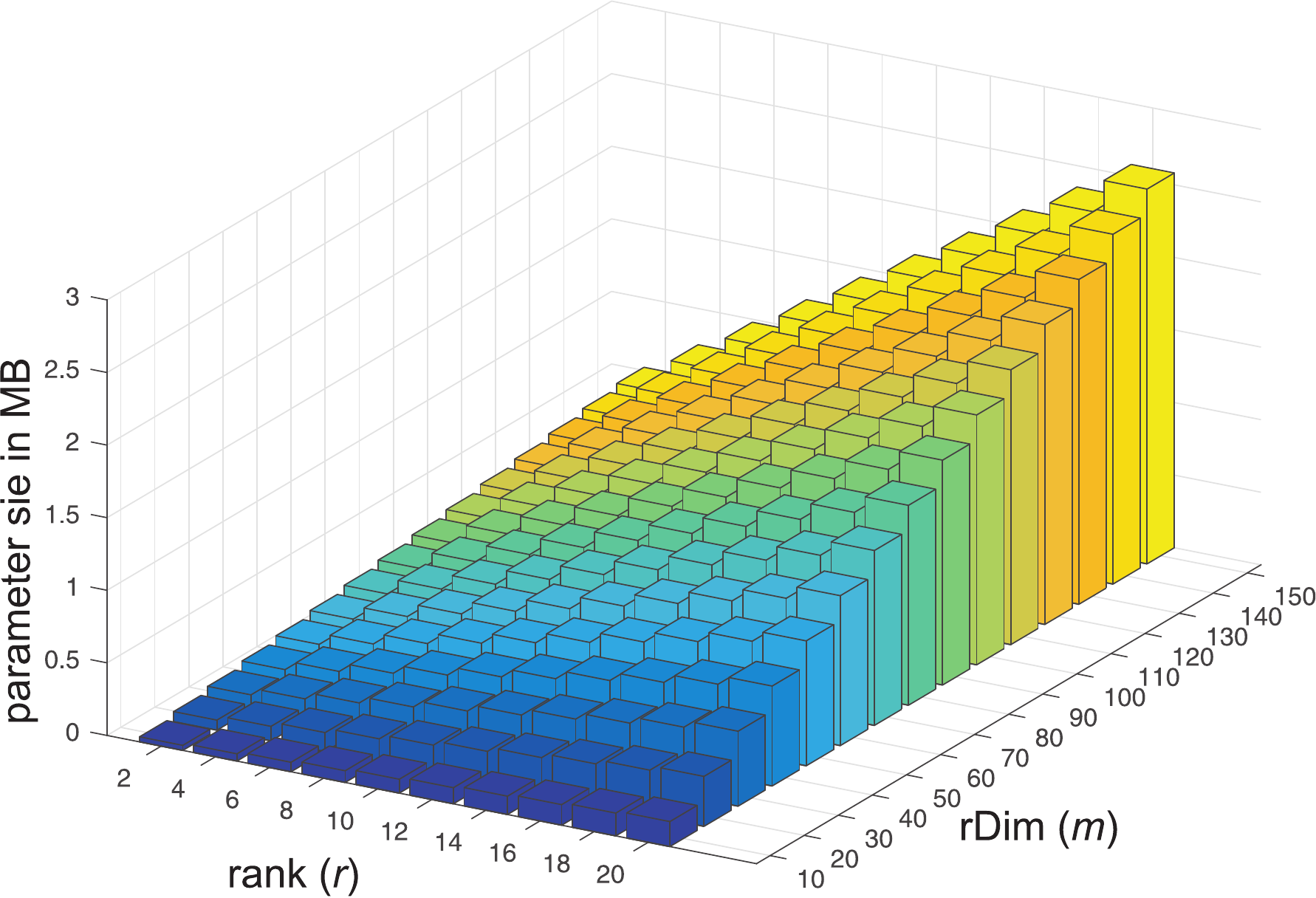}
        \captionsetup{width=0.9\textwidth}
        \caption{The learned parameter size versus reduced dimension ($m$) and rank ($r$).}
        \label{fig:paramSize_dimRed_rank}
    \end{minipage}
\end{figure*}

\subsection{Configuration of Hyperparameters}
\label{ssec:configuration}
Two hyperparameters are involved in specifying our architecture, the dimension
$m$ in the subspace determined by ${\bf P} \in \RB^{c \times m}$ and the rank
$r$ of the classifiers $\V_k \in \RB^{m\times r}$ for $k=1,\dots, K$.
To investigate these two parameters in our model, we conduct an experiment on
CUB-200-2011 bird dataset~\cite{wah2011caltech}, which contains $11,788$ images
of 200 bird species, with a standard training and testing set split.  We do not
use any part annotation or masks provided in the dataset.

We first train a full-rank model on the bilinear pooled features and then
decompose each classifier using eigenvalue decomposition and keep the largest
magnitude eigenvalues and the corresponding vectors to produce a rank-$r$
classifier.  After obtaining low-rank classifiers, we apply co-decomposition as
described in Section~\ref{sec:co-decomposition} to obtain projector $\bf P$ and
compact classifiers $\V_k$'s. We did not perform fine-tuning of these models
but this quick experiment provides a good proxy for final model performance
over a range of architectures.

We plot the classification accuracy vs. rank $r$ and reduced dimension $m$
(rDim) in Figure~\ref{fig:acc_dimRed_rank}, the average reconstruction fidelity
measured by peak signal-to-noise ratio to
the original classifier parameters $\U_k$ versus rank $r$ and dimension $m$ in
Figure~\ref{fig:reconErr_vs_rDimRank_lsvm}, and model size versus rank $r$ and
dimension $m$ in Figure~\ref{fig:paramSize_dimRed_rank}.

As can be seen, the reconstruction fidelity (measured in the peak signal-to-noise ratio)
is a good guide to model performance prior to fine tuning.  Perhaps
surprisingly, even with $r=8$ and $m=100$, our model achieves near-maximum
classification accuracy on this dataset (Figure \ref{fig:acc_dimRed_rank}) with
model parameters compressed by a factor of 100 over the full model
(Figure \ref{fig:paramSize_dimRed_rank}).  Based on this analysis, we set $r=8$
and $m=100$ for our quantitative benchmark experiments.

\subsection{Baseline Methods}
We use VGG16~\cite{simonyan2014very} as the base model in all comparison to be
consistent with previous work~\cite{lin2015bilinear,gao2015compact}.

\paragraph{Fully Connected layers (FC-VGG16):}
We replace the last fully connected layer of VGG16 base model with a randomly
initialized $K$-way classification layer and fine-tune.  We refer this as
``FC-VGG16'' which is commonly a strong baseline for a variety of computer
vision tasks.  As VGG16 only takes input image of size $224\times 224$, we
resize all inputs for this method.

\paragraph{Improved Fisher Encoding (Fisher):}
Fisher encoding~\cite{perronnin2010improving} has recently been used as an
encoding and pooling alternative to the fully connected
layers~\cite{cimpoi2015deep}. Consistent
with~\cite{gao2015compact,lin2015bilinear}, we use the activations at layer
$conv5\_3$ (prior to ReLU) as local features and set the encoding to use 64 GMM
components for the Fisher vector representation.

\paragraph{Full Bilinear Pooling (Full Bilinear):}
We use full bilinear pooling over the $conv5\_3$ feature maps (termed
``symmetric structure'' in \cite{lin2015bilinear}) and apply element-wise sign
square root normalization and $\ell_2$ normalization prior to classification.

\paragraph{Compact Bilinear Pooling:}
We report two methods proposed in~\cite{gao2015compact} using Random Maclaurin
and Tensor Sketch.  Like Full Bilinear model, element-wise sign square root
normalization and $\ell_2$ normalization are used.  We set the projection
dimension $d=8,192$, which is shown to be sufficient for reaching close-to
maximum accuracy~\cite{gao2015compact}.  For some datasets, we use the code
released by the authors to train the model; otherwise we display the
performance reported in~\cite{gao2015compact}.

\begin{table*}
\centering
\caption{Classification accuracy and parameter size of: a fully connected
network over VGG16~\cite{simonyan2014very}, Fisher
vector~\cite{cimpoi2015deep}, Full bilinear CNN~\cite{lin2015bilinear}, Random
Maclaurin~\cite{gao2015compact}, Tensor Sketch~\cite{gao2015compact}, and our
method.  We run Random Maclaurin and Tensor Sketch with the code provided
in~\cite{gao2015compact} with their conventional configuration (\eg projection
dimension $d=8192$).
}
\begin{tabular}{l | c c c c c c }
\hline
            & FC-VGG16   & Fisher        & Full Bilinear & Random Maclaurin  & Tensor Sketch  & LRBP (Ours)  \\
\hline
CUB~\cite{wah2011caltech}
            & 70.40      & 74.7          & 84.01             & 83.86      &   84.00           & \textbf{84.21}\\
DTD~\cite{cimpoi2014describing}
            & 59.89      & 65.53         & 64.96             & 65.57      &   64.51           & \textbf{65.80}\\
Car~\cite{krause20133d}
            & 76.80      & 85.70         & 91.18             & 89.54      &   90.19           & \textbf{90.92} \\
Airplane~\cite{maji2013fine}
            & 74.10      & 77.60         & 87.09             & 87.10      &   87.18           & \textbf{87.31} \\
\hline
param. size (CUB)
            & 67MB      & 50MB         & 200MB             & 48MB      &   8MB           & 0.8MB\\
\hline
\end{tabular}
\label{tab:clsResults}
\end{table*}

\begin{table}
\centering
\caption{Summary statistics of datasets.}
\begin{tabular}{l | c c c  }
\hline
            & $\#$ train img. & $\#$ test img. & $\#$ class   \\
\hline
CUB~\cite{wah2011caltech}
            & 5994      & 5794 & 200 \\
DTD~\cite{cimpoi2014describing}
            & 1880      & 3760 & 47  \\
Car~\cite{krause20133d}
            & 8144      & 8041 & 196 \\
Airplane~\cite{maji2013fine}
            & 6667      & 3333 & 100 \\
\hline
\end{tabular}
\label{tab:dataset}
\end{table}

\subsection{Quantitative Benchmarking Experiment}
\label{ssec:exp}

We compare state-the-art methods on four widely used fine-grained
classification benchmark datasets, CUB-200-2011 Bird
dataset~\cite{wah2011caltech}, Aircrafts~\cite{maji2013fine},
Cars~\cite{krause20133d}, and describing texture dataset
(DTD)~\cite{cimpoi2014describing}.  All these datasets provide fixed train and
test split.  We summarize the statistics of datasets in
Table~\ref{tab:dataset}.  In training all models, we only use the category
label without any part or bounding box annotation provided by the datasets.  We
list the performance of these methods in Table~\ref{tab:clsResults} and
highlight the parameter size of the models trained on CUB-200 dataset in the
last row.

From the comparison, we can clearly see that Fisher vector pooling not only
provides a smaller model than FC-VGG16, but also consistently outperforms it by
a notable margin.  All the bilinear pooling methods, including ours, achieve
similar classification accuracy, outperforming Fisher vector pooling by a
significant margin on these datasets except DTD.  However, our model is
substantially more compact than the other methods based on bilinear features.
To the best of our knowledge, our model achieves the state-of-the-art
performance on these datasets without part
annotation~\cite{jaderberg2015spatial, krause2015fine}, and even outperforms
several recently proposed methods trained that use supervised part
annotation~\cite{zhang2015fine}.  Although there are more sophisticated methods
in literature using detailed annotations such as parts or bounding
box~\cite{zhang2014part,zhang2016}, our model relies only on the category
label.  These advantages make our model appealing not only for
memory-constrained devices, but also in weakly supervised fine-grained
classification in which detailed part annotations are costly to obtain while
images with category label are nearly free and computation during model
training becomes the limiting resource.

\subsection{Qualitative Visualization}
\label{sec:Qualitative}

\begin{figure*}[t]
\centering
   \includegraphics[width=1\linewidth]{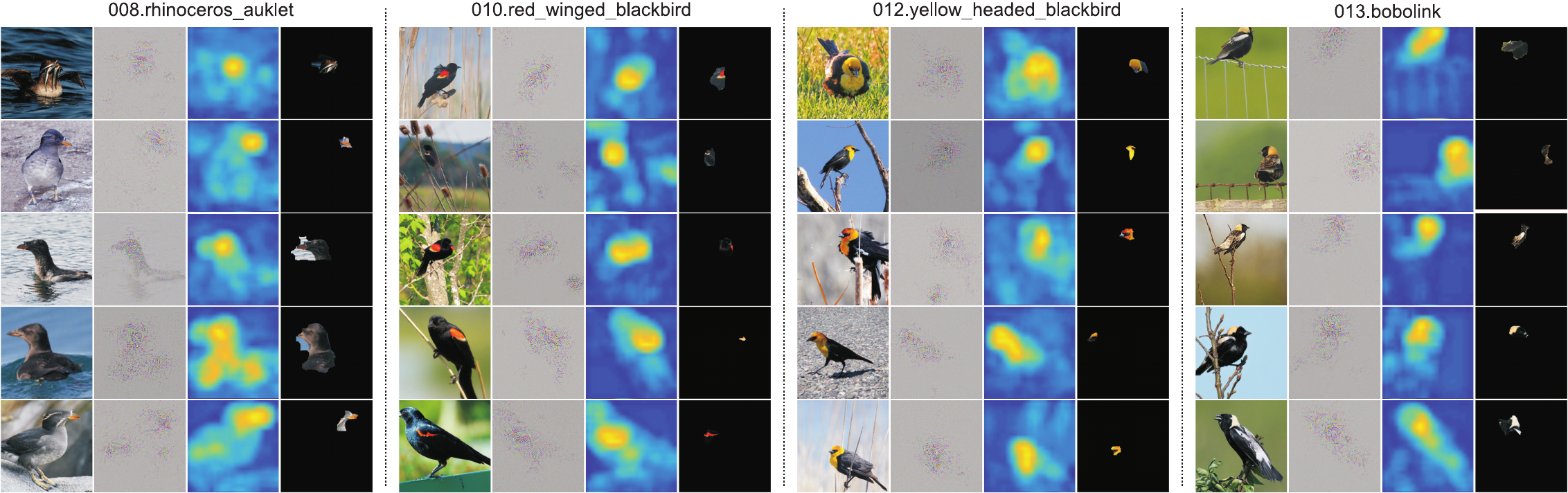}
   \caption{(Best seen in color.)
   In each panel depicting a different bird species, the four columns show the
   input images and the visualization maps using three different methods as
   described in Section~\ref{sec:Qualitative}.  We can see our model tends to
   ignore features in the cluttered background and focus on the most distinct
   parts of the birds.
   }
\label{fig:focusHighlight}
\end{figure*}

To better understand our model, we adopt three different approaches to
visualizing the model response for specific input images.  In the first method,
we feed an input image to the trained model, and compute responses $\Y =
[{\U_+}_1, {\U_-}_1, \dots, {\U_+}_k, {\U_-}_k, \dots, {\U_+}_K, {\U_-}_K]^T
\X$ from the bilinear classifier layer.  Based on the ground-truth class label,
we create a modified response $\bar\Y$ by zeroing out the part corresponding to
negative Frobenius score ($-\Vert\U_-^T\X\Vert_F^2$) for the ground-truth
class, and the part to the positive Frobenius scores ($\Vert\U_+^T\X\Vert_F^2$)
in the remaining classifiers, respectively. This is similar to approaches used
for visualizing HOG templates by separating the positive and negative
components of the weight vector. To visualize the result, we treat $\bar\Y$ as
the target and backpropagate the difference to the input image space, similar
to~\cite{simonyan2013deep}.  For the second visualization, we compute the
magnitude of feature activations averaged across feature channels used by the
bilinear classifier.  Finally, we produce a third visualization by repeatedly
remove superpixels from the input image, selecting the one that introduces
minimum drop in classification score This is similar
to~\cite{ribeiro2016should,zhou2014learning}.  In
Figure~\ref{fig:focusHighlight}, we show some randomly selected images from
four different classes in CUB-200-2011 dataset and their corresponding
visualizations.

The visualizations all suggest that the model is capable of ignoring cluttered
backgrounds and focuses primarily on the bird and even on specific
discriminative parts of each bird.  Moreover, the highlighted activation region
changes w.r.t the bird size and context, as shown in the first panel of
Figure~\ref{fig:focusHighlight}.  For the species
``010.red\_winged\_blackbird'', ``012.yellow\_headed\_blackbird'' and
``013.bobolink'', the most distinctive parts, intuitively, are the red wings,
yellow head and neck, and yellow nape, respectively. Our model naturally
appears to respond to and localize these parts. This provides a partial
explanation as to why simple global pooling achieves such good results without
an explicit spatial transformer or cross-channel pooling architecture
(e.g.~\cite{liu2015treasure})

\section{Conclusion}
We have presented an approach for training a very compact low-rank
classification model that is able to leverage bilinear feature pooling for
fine-grained classification while avoiding the explicit computation of
high-dimensional bilinear pooled features.  Our Frobenius norm based classifier
allows for fast evaluation at test time and makes it easy to impose hard,
low-rank constraints during training, reducing the degrees of freedom in the
parameters to be learned and yielding an extremely compact feature set.  The
addition of a co-decomposition step projects features into a shared subspace
and yields a further reduction in computation and parameter storage. Our
final model can be initialized with a simple PCA step followed by end-to-end
fine tuning.

Our final classifier model is one to two orders of magnitude smaller than
existing approaches and achieves state-of-the-art performance on several public
datasets for fine-grained classification by using only the category label
(without any keypoint or part annotations). We expect these results will
form a basis for future experiments such as training on weakly supervised
web-scale datasets~\cite{krause2015unreasonable}, pooling multiple feature
modalities and further compression of models for use on mobile devices.

\section*{Acknowledgement}
This project is supported by NSF grants
IIS-1618806, IIS-1253538, DBI-1262547 and a hardware donation
from NVIDIA.

{\small
\bibliographystyle{ieee}
\bibliography{egbib}
}

\end{document}